\documentclass{article}

\usepackage[final]{neurips_mlsb_2022}

\usepackage[utf8]{inputenc} %
\usepackage[T1]{fontenc}    %
\usepackage{hyperref}       %
\usepackage{url}            %
\usepackage{booktabs}       %
\usepackage{amsfonts}       %
\usepackage{nicefrac}       %
\usepackage{microtype}      %
\usepackage{xcolor}         %
\usepackage{amsmath}
\usepackage{mathtools}
\usepackage{enumitem}

\newcommand{\wt}{x^{\text{wt}}}
\newcommand{\mt}{x^{\text{mt}}}
\newcommand{\nvariants}{10011}
\newcommand{\ngenes}{1348}

\title{Unsupervised language models for disease variant prediction}

\author{%
    Allan Zhou\thanks{Equal contribution.} \\
    Stanford University
    \And Nicholas C. Landolfi$^*$ \\
    Stanford University
    \And Daniel C. O'Neill \\
    Stanford University
}

\begin{document}

\maketitle

\begin{abstract}

There is considerable interest in predicting the pathogenicity of protein variants in human genes. Due to the sparsity of high quality labels, recent approaches turn to \textit{unsupervised} learning, using Multiple Sequence Alignments (MSAs) to train generative models of natural sequence variation within each gene. These generative models then predict variant likelihood as a proxy to evolutionary fitness. In this work we instead combine this evolutionary principle with pretrained protein language models (LMs), which have already shown promising results in predicting protein structure and function. Instead of training separate models per-gene, we find that a single protein LM trained on broad sequence datasets can score pathogenicity for any gene variant zero-shot, without MSAs or finetuning. We call this unsupervised approach \textbf{VELM} (Variant Effect via Language Models), and show that it achieves scoring performance comparable to the state of the art when evaluated on clinically labeled variants of disease-related genes.
\end{abstract}

\section{Introduction}

Understanding and quantifying the pathogenicity of human gene variants could transform healthcare, better inform treatment decisions, and enable new treatment modalities. 
However, relating specific missense variants to phenotypical disease indications is challenging, since the number of such variants (6.5 million) observed in the human population so far exceeds that which can be analyzed \cite{karczewski2020mutational}. 
Despite large-scale efforts to collate the disease relevance of gene variants \cite{landrum2018clinvar}, the majority of variants remain pathogenically unclassified \cite{van2020exome}.

Computational methods offer the promise of at-scale interpretation of variants at speeds useful in a clinical setting \cite{jagadeesh2019s,rentzsch2019cadd}. However, many supervised models are trained on clinical labels of variable quality or with inconsistent clinical annotations resulting in inconsistent model performance. Unsupervised generative models avoid the labeling issues and have been successfully used to predict protein function and stability \cite{hopf2014sequence,lapedes2012using,meier2021language}. More recently, \citet{frazer2021disease} introduced EVE, a family of variational autoencoders (VAEs) trained on protein Multiple Sequence Alignments (MSAs) for each gene of interest. EVE scores pathogenicity using variant probabilities as proxies for evolutionary fitness, and achieves current state-of-the art performance compared to other computational approaches without training on clinical labels.

In this work we describe \textbf{VELM} (Variant Effect via Language Models), an unsupervised approach for scoring variant pathogenicity using protein language models (LMs). Like prior unsupervised evolutionary approaches, VELM scores pathogenicity by using a sequence model to predict sequence likelihood. However, instead of training separate gene-specific generative models to estimate likelihoods, we use protein LMs pretrained by self-supervised learning on large open datasets of protein sequences. This training procedure produces models that capture statistical patterns across a broad distribution of protein sequences, and enables estimating sequence likelihood for any gene variant zero-shot without finetuning on any MSAs. Thus, our approach uses a single model to efficiently scores pathogenicity for any gene variant of interest without having to train a new generative model per-gene. Ultimately, VELM allows us to efficiently predict pathogenicity for the large number of currently unlabeled variants across human disease-related genes.

\begin{figure}
\centering
\includegraphics[width=0.48\textwidth]{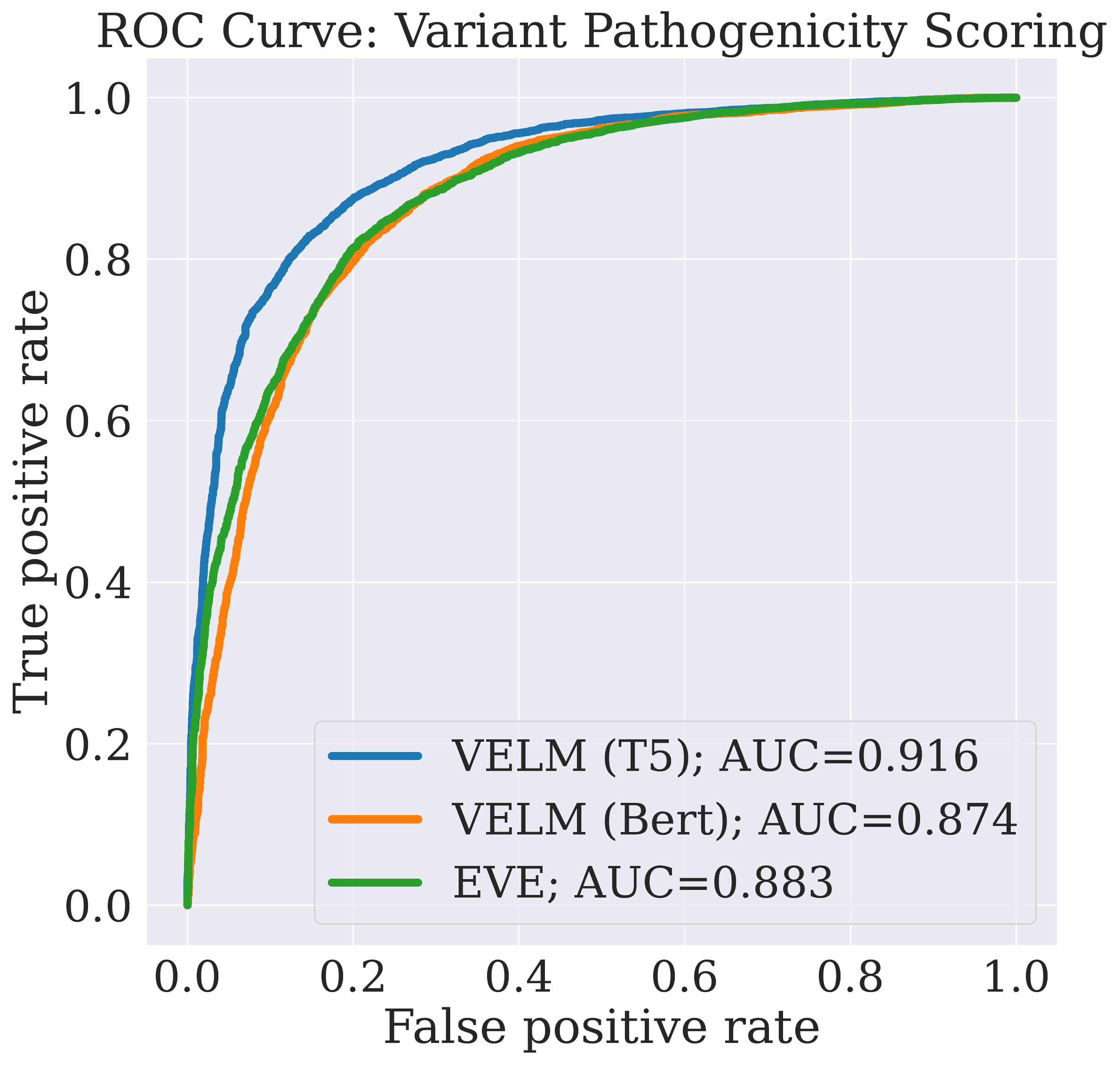}
\raisebox{0.15\height}{\includegraphics[width=0.48\textwidth]{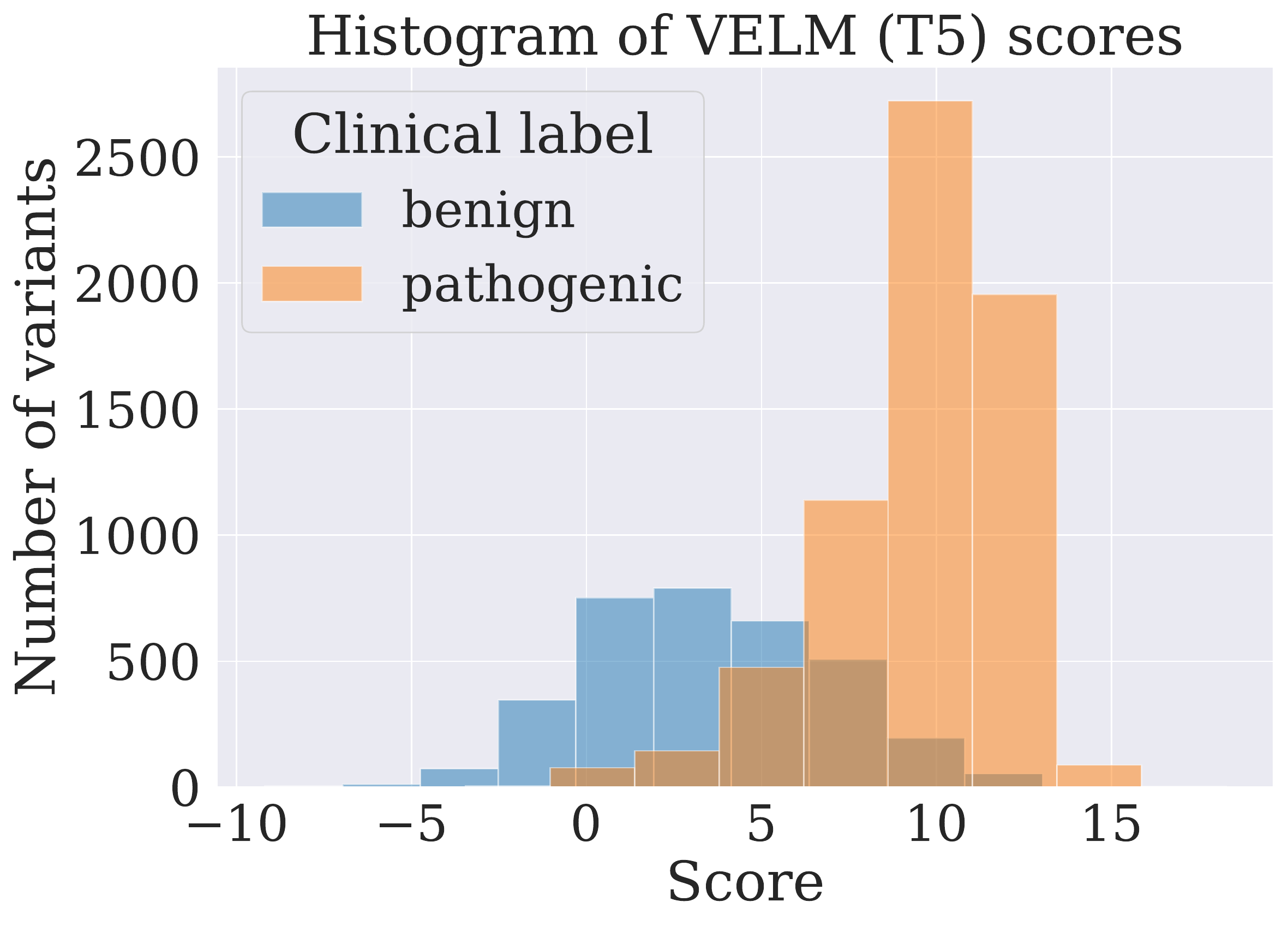}}
\caption{Left: Receiver Operating Characteristic (ROC) curve of VELM and EVE scores on our evaluation set of clinically labeled gene variants. VELM (T5) outperforms both the EVE score and VELM (Bert). Positive $\iff$ pathogenic, negative $\iff$ benign. Right: Histogram of VELM (T5) scores on clinically labeled variants. Broadly speaking, VELM assigns higher scores to pathogenic variants than for benign ones.}
\label{fig:aggregate-roc}
\end{figure}

When evaluated on a set of variants with known clinical labels from the ClinVar dataset~\citep{landrum2018clinvar}, we find that the VELM score can discriminate variant pathogenicity with an AUC=$0.92$, exceeding the performance of EVE (AUC=$0.89$), see Figure~\ref{fig:aggregate-roc}.

\section{VELM: Variant Effect via Language Models}

Starting from a reference wildtype protein, our goal is to predict the pathogenicity of a given variant directly from its protein sequence. Following an unsupervised evolutionary approach, we leverage the relationship between sequence likelihood and evolutionary fitness to score variants without training on clinical labels (which can cause overfitting). We estimate sequence likelihood through protein language models (LMs) pretrained on large protein sequence datasets. Compared to EVE~\citep{frazer2021disease}, this removes the need to train separate per-gene generative models on processed MSAs. Indeed, we will show that a single pretrained LM can score any gene variant with no finetuning.

Inspired by techniques from natural language processing (NLP), protein language models are typically trained on large datasets of protein sequences with a masked language modeling objective. This trains the model to estimate the distribution over residues at particular positions given the \textit{context} residues at surrounding positions. More precisely, these models compute $P(x_{i_1} = \bullet,\cdots,x_{i_m}=\bullet | x_{\backslash \{i_1,\cdots,i_m\}})$, where in practice the context $x_{\backslash \{i_1,\cdots,i_m\}}$ is created by masking the sequence at positions $i_1, \cdots, i_m$. 

To define the VELM pathogenicity score, we need to use the protein LM to estimate a notion of variant likelihood (relative to the wildtype). We denote the wildtype sequence $\wt$ and variant sequence $\mt$, and define the set of mutation positions $M = \{i: \mt_i \neq \wt_i\}$. \citet{meier2021language} found that the log odds ratio at mutated positions can effectively predict protein function. We define the VELM score using the same approach:
\begin{equation}
\label{eq:score}
S(\mt) \vcentcolon = \sum_{i \in M} \log P(x_i = \wt_i | \wt_{\backslash M}) - \log P(x_i = \mt_i | \mt_{\backslash M})
\end{equation}
where $x_{\backslash M}$ indicates masking $x$ at all positions $i\in M$ (notably, $\mt_{\backslash M} = \wt_{\backslash M}$). Intuitively, $S(\mt)$ should be higher when the variant is \textit{less} likely, indicating that it is more likely to be pathogenic. Computing $S(\mt)$ is relatively efficient and requires $|M|$ forward passes to evaluate a single variant. For reasonably small $|M|$, GPU batching leads to only a single forward pass in practice.

\section{Experiments and Analysis}

We apply VELM to missense variants of human disease-related genes whose sequence lengths are $\leq 512$\footnote{This is not a general limitation of VELM, but the particular protein LMs we use in this evaluation were only trained on sequences of length $\leq 512$.}. From the ClinVar dataset~\citep{landrum2018clinvar} there are known clinical labels for \nvariants{} variants across the \ngenes{} genes we consider\footnote{We restrict to those labels with a ClinVar quality rating of at least one star.
For comparison purposes, we only consider variants involving a mutation at an EVE \emph{focus} position. 
Unlike VELM, EVE only scores mutations at positions with sufficient MSA coverage.}: 6613 variants are labeled ``pathogenic'' while 3398 are labeled ``benign.'' For these clinically labeled variants, we compare our VELM score against the value of the label, and evaluate the effectiveness of using VELM score to classify variant pathogenicity.

Computing the VELM pathogenicity score (Eq.~\ref{eq:score}) requires a pretrained protein LM, for which there are multiple choices. Here we consider both ProtBert (420M parameters) and ProtT5 (3B parameters)~\citep{elnaggar2021prottrans}, both trained by masked language modeling on BFD~\citep{steinegger2018clustering} and UniRef~\citep{suzek2015uniref}. We will denote the results of scoring variants with each LM as \textbf{VELM (Bert)} and \textbf{VELM (T5)}, respectively. For comparison, we also evaluate the performance of other methods on the same set of variants:
\begin{enumerate}[wide, labelwidth=!, labelindent=0pt]
\item EVE~\citep{frazer2021disease}: An unsupervised evolutionary method that trains separate generative models on MSAs for each gene.
\item MutationAssessor (MA)~\cite{reva2011predicting}: Another unsupervised scoring approach.
\item DEOGEN2 (DG2)~\citep{raimondi2017deogen2}: A supervised method trained on clinical disease labels.
\item REVEL~\citep{ioannidis2016revel}: An ensemble method that combines the output of multiple individual tools.
\end{enumerate}

\begin{table}
\begin{center}
\begin{tabular}{ c|c|c|c|c|c|c }
\hline
 Method & \small{VELM~(T5)} & \small{VELM~(Bert)} & EVE & REVEL & MA & DG2 \\ 
 \hline
 mAUC ($\geq 1$ labels) & 0.901 & 0.858 & 0.917 & 0.934 & 0.888 & 0.895 \\
 mAUC ($\geq 3$ labels) & 0.912 & 0.876 & 0.930 & 0.946 & 0.895 & 0.901 \\
 mAUC ($\geq 5$ labels) & 0.933 & 0.892 & 0.936 & 0.956 & 0.904 & 0.916 \\
 \hline
\end{tabular}
\caption{Mean of AUCs (mAUC) over the evaluation set of disease-relevant genes (weighted by number of known labels). For each row, ``$\geq N$ labels'' means we restrict evaluation to genes that have at least $N$ pathogenic and $N$ benign labels for evaluating AUC. Note that VELM (ours), EVE~\citep{frazer2021disease} and MA (MutationAssessor) are all unsupervised methods. DG2 (DEOGEN2)~\citep{raimondi2017deogen2} is supervised by clinical labels, while REVEL~\citep{ioannidis2016revel} is an ensemble method that combines the output of multiple individual tools.}
\label{table:mean-auc}
\end{center}
\vspace{-1.5em}
\end{table}

\textbf{Aggregate Metrics}: Figure~\ref{fig:aggregate-roc} shows how the VELM score discriminates pathogenicity on our set of labeled variants. The VELM~(T5) score has an AUC of 0.92, outperforming both EVE and VELM~(Bert), with AUCs 0.88 and 0.87, respectively. The ROC Curve indicates that VELM~(T5) produces pathogenicity scores with an overall better tradeoff beteween TPR and FPR compared to the other methods. The histogram of scores shaded by clinical label shows that the VELM scores is broadly capable of separating pathogenic and benign gene variants. Since the score is simply computed from the output of a protein LM, this indicates that the pretraining process learns statistical patterns in protein sequences that are relevant to predicting pathogenicity (via predicting likelihood).

\textbf{Per-Gene Metrics}: We can also evaluate how VELM scores discriminate pathogenicity on a per-gene basis. We calculate the \textit{Mean AUC} (mAUC) by computing AUC for variants of each gene separately, then average the AUCs over genes weighted by the number of clinical labels available. Since many genes have just a few clinically labeled variants, per-gene evaluation statistics may be very noisy. We separately evaluate mAUC over genes with at least $N$ pathogenic and benign labels, where $N=1, 3, $ or $5$. Table~\ref{table:mean-auc} shows that REVEL generally achieves the highest Mean AUC on each evaluation set. Among non-ensemble methods, EVE generally performs best, though for the least noisy evaluation set of genes with $\geq 5$ labels, VELM~(T5) and EVE perform comparably.

\subsection{Analysis}
Overall, VELM~(T5) achieves state of the art performance at predicting pathogenicity for arbitrary gene variants (aggregate AUC). It is comparable to other methods when scoring at a per-gene level (mean AUC), nearly matching state of the art for the least noisy evaluation set. These results are notable since VELM simply uses a pretrained protein LM to score any gene variant zero-shot, while other methods either train on clinical labels or on gene-specific evolutionary data. This leaves open the possibility for further improving performance by finetuning the protein LM on data pertaining to the disease-relevant genes of interest.

The fact that VELM (T5) outperforms VELM (Bert) falls in line with prior observations that ProtT5 outperforms ProtBert on a variety of structure and function prediction tasks~\cite{elnaggar2021prottrans}. This suggests that pathogenicity prediction may be yet another ``downstream task'' where performance can improved by simply pretraining better protein LMs.

\section{Related Work}

There has been extensive prior work in computational techniques to predict protein pathogenicity and in using large-scale self-supervised language models for protein sequences.

The literature on computational approaches for predicting protein pathogenicity is large and growing.
Roughly speaking, these approaches can be categorized into \textit{supervised} methods (e.g., \cite{adzhubei2010method, raimondi2017deogen2}), \textit{unsupervised methods} (e.g., \cite{sim2012sift, choi2012predicting}), and supervised \textit{meta-predictor} methods that use the outputs of both supervised and unsupervised methods as features (e.g.,  \cite{ioannidis2016revel,jagadeesh2016m,feng2017perch, qi2021mvp, ionita2016spectral}).
The unsupervised approach is favored in prior work which cites the variable quality of labels, bias in label availability, and sparsity of labels as difficulties in developing and validating supervised methods.
In comparing to our work, the most recent and relevant such \textit{unsupervised} approach is EVE \cite{frazer2021disease}, which is state-of-the-art.
The key features distinguishing our work from that of \cite{frazer2021disease} are: (a) we have one global protein LM instead of per-family sequence models (b) we train on a large database of protein sequences with no fine-tuning instead of EVE's individual MSAs, and (c) we perform zero-shot inference across all residue locations of a protein, instead of EVE's \textit{focus} indices.

There has been a recent growth of interest in training language models on protein sequence datasets for the purposes of predicting protein structure and function~\citep{alley2019unified,lu2020self,madani2020progen,elnaggar2021prottrans,rives2021biological,notin2022tranception}. Most closely related to our work is ESM-1v~\citep{meier2021language}, which used protein LMs and the log odds ratio at mutated positions to predict the effect of mutations on protein function zero-shot. Given the success of protein LMs for predicting structure and function, VELM explores their effectiveness for directly predicting pathogenicity in disease-relevant human genes.

In concurrent work, \citet{brandes2022genome} also score variant pathogenicity zero-shot with a similar approach but using the ESM1b~\cite{rives2021biological} model. Their more extensive analysis corroborates our preliminary findings: in particular, the observation that protein LMs outperform EVE on aggregate metrics but not on per-gene metrics.

\section{Conclusion}
In this work, we investigate the effectiveness of pretrained protein language models for assessing variant pathogenicity, a problem of great clinical interest. We introduce an unsupervised method called VELM that scores variant sequences by using protein LMs to estimate sequence likelihood, and show that it matches state of the art predictive performance. VELM is computationally efficient and flexible, using a single model to score variants of any gene with no finetuning.

The current work can be improved along multiple directions. First, the current protein LMs were trained on sequences of limited length, restricting our evaluation to sequences of length $\leq 512$. Aside from removing this technical limitation, results can likely be improved by using better pretrained LMs such as ESM~\citep{rives2021biological}, or by finetuning the LMs on relevant sequences (to human disease-related genes).

\bibliographystyle{abbrvnat}
\bibliography{references}

\end{document}